\documentclass[journal]{IEEEtran}
\ifCLASSINFOpdf
\else
\fi

\usepackage{epsfig}
\usepackage{cite}
\usepackage{graphicx}
\usepackage{amsmath}
\usepackage{amssymb}
\usepackage{enumerate}
\usepackage{subcaption}
\usepackage{multirow}
\usepackage{array}
\usepackage{algorithm}
\usepackage{algorithmic}
\usepackage{epstopdf}
\usepackage{amsthm}
\usepackage{float}
\usepackage{array}
\usepackage{bm}
\usepackage{color}
\usepackage{latexsym}
\begin{document}
%
\title{SiGAN: Siamese Generative Adversarial Network for Identity-Preserving Face Hallucination}
%
%
%

	\author{Chih-Chung~Hsu,~\IEEEmembership{Member,~IEEE,}
	Chia-Wen~Lin,~\IEEEmembership{Fellow,~IEEE,}
	Weng-Tai~Su,~\IEEEmembership{Student~Member,~IEEE,}
	and~Gene~Cheung,~\IEEEmembership{Senior~Member,~IEEE,}
\thanks{Manuscript received June 20, 2018. This paper was supported in part by the Ministry of Science and Technology, Taiwan, under Grants MOST 106-2221-E-007-079 -MY3.}
\thanks{Chih-Chung Hsu is with Department of Management Information Systems, National Pingtung University of Science and Technology, Pingtung, Taiwan. (e-mail: cchsu@mail.npust.edu.tw)}
\thanks{Chia-Wen Lin (corresponding author) is with the Department of Electrical Engineering and the Institute of Communications Engineering, National Tsing Hua University, Hsinchu, Taiwan. (e-mail: cwlin@ee.nthu.edu.tw)}
\thanks{Weng-Tai Su is with the  Department of Electrical Engineering, National Tsing Hua University, Hsinchu, Taiwan.}
\thanks{Gene Cheung is with the National Institute of Informatics, Tokyo, Japan. (e-mail: cheung@nii.ac.jp)}
\thanks{Color versions of one or more of the figures in this paper are available online at http://ieeexplore.ieee.org.}}

\maketitle

\begin{abstract}
Despite generative adversarial networks (GANs) can hallucinate photo-realistic high-resolution (HR) faces from low-resolution (LR) faces,  they cannot guarantee preserving the identities of hallucinated HR faces, making the HR faces poorly recognizable. To address this problem, we propose a Siamese GAN (SiGAN) to reconstruct HR faces that visually resemble their corresponding identities. On top of a Siamese network, the proposed SiGAN consists of a pair of two identical generators and one discriminator.  We incorporate reconstruction error and identity label information in the loss function of SiGAN in a pairwise manner. By iteratively optimizing the loss functions of the generator pair and discriminator of SiGAN, we cannot only achieve photo-realistic face reconstruction, but also ensures the reconstructed information is useful for identity recognition. Experimental results demonstrate that SiGAN significantly outperforms existing face hallucination GANs in objective face verification performance, while achieving photo-realistic reconstruction. Moreover, for input LR faces from unknown identities who are not included in training, SiGAN can still do a good job.
\end{abstract}

\begin{IEEEkeywords}
Face hallucination, convolutional neural networks, generative adversarial networks, super-resolution, generative model.
\end{IEEEkeywords}

%
\IEEEpeerreviewmaketitle

\section{Introduction}
\label{sec:intro}
\IEEEPARstart{F}{ace} hallucination that super-resolves a low-resolution (LR) face image to a high-resolution (HR) one has become an attractive technique in upscaling face photos because it has found many applications such as security in surveillance video, face recognition, face tracking, facial expression estimation, etc. which usually require face images with enough fine details. However, simple interpolation schemes cannot reconstruct fine details. Instead, example-based super-resolution (SR) schemes \cite{baker2002limits} have proven to be able to reconstruct significantly finer details from a LR image compared to interpolation-based schemes, provided that a comprehensive set of training HR/LR image pairs is used to learn the structures and patterns of face image pairs based on machine learning techniques.

The problem with face hallucination is, however, different from that with generic image SR because face images have unified structures which people are very familiar with. Even only few reconstruction errors occurring on a face will cause visually annoying artifacts. For example, geometry distortion in the mouth and eyes on a reconstructed face may only slightly reduce the objective quality of image, but can significantly  hurt the perceived quality subjectively. Therefore, both the global face shape and textures and local geometric structures (e.g., mouth, nose, and eyes) need to be treated carefully in face hallucination \cite{baker2000hallucinating}\cite{wang2014comprehensive}.

\begin{figure}
	\centering
	\includegraphics[width=0.45\textwidth]{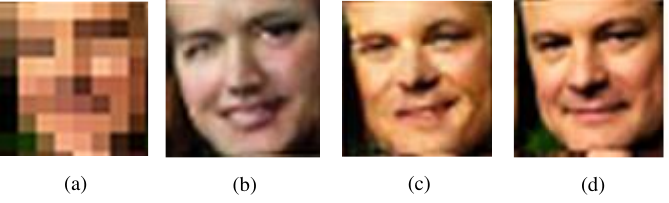}
	\caption{\small Illustration of face hallucination: (a) input LR face ($8 \times 8$); HR faces reconstructed by (b) identity-unaware face hallucination; (c) identity-aware face hallucination (our method); and (d) the ground-truth.}
	\label{fig:header_fig}	
\end{figure}

To recognize the identity of a LR face captured by a surveillance camera
is a challenging problem, as face images are often taken in a distance, making their resolutions too low to provide sufficiently discriminative features. Recently, empirical studies \cite{zou2012very} in face recognition revealed that a minimum face resolution between $32\times 32$ and $64\times 64$ is required for effective face recognition, and an even lower resolution would degrade recognition performance significantly for existing recognition models. It is therefore desirable to develop an effective face hallucination scheme.

 Most of existing face hallucination methods \cite{wang2014comprehensive}, nevertheless, were just focused on hallucinating visually pleasant HR details without considering whether the added details are helpful in recognizing the identity of a face. As illustrated in Fig. \ref{fig:header_fig}(b), such identity-unaware reconstructed faces, though with a higher resolution, usually cannot help boost face recognition/verification accuracy. Instead, identity-aware face hallucination, that can hallucinate identity-preserving facial details as shown in Fig. \ref{fig:header_fig}(c),  much better serves the purpose. Identity-preserving reconstruction is therefore vital in face hallucination for many real-world applications \cite{jfhfr2008,wu2016deep}. 
 
  Hallucinating identify-preserving HR faces requires a labeled training set to learn identity-preserving representations. Nevertheless, since collecting a large-scale well-labeled face dataset is very costly, it is therefore desirable to develop a learning methodology that can deal with weakly-labeled training dataset to drastically reduce the labeling cost.  
 
 To address the above problems, in this paper, we propose a novel Siamese generative adversarial network (SiGAN) to hallucinate HR faces to achieve photo-realistic and identity-preserving reconstruction. The training of proposed SiGAN only relies on weak pairwise labels, that signify whether a pair of two faces belongs to a same identity without the need of knowing the true identities of faces, thanks to its Siamese network structure. Our contributions are summarized below:
 
 \begin{itemize}
 	\item We propose a novel face hallucination GAN on top of a Siamese Network (namely SiGAN), upon which we can hallucinate HR faces to achieve photo-realistic and identity-preserving reconstruction.
 	\item We embed weak binary pairwise label information by a Siamese network without the need of true labels, which significantly reduces the labeling cost and increases the scalability of the method for faces belonging to unseen identities.
 	\item Perceptive and quantitative experiments demonstrate the outstanding performance and considerable generalization ability of the proposed SiGAN.
 \end{itemize}

The rest of this paper is organized as follows.
Some most relevant works are surveyed in Sec. \ref{sec:related}. Sec. \ref{sec:sigan} presents the proposed Siamese GAN for identity-aware face hallucinations. To compare with the proposed SiGAN, we also present two other implementations of identity-embedding face hallucination GANs in Sec. \ref{sec:fhgan}. In Sec. \ref{sec:experiments}, experimental results are demonstrated. Finally, conclusions are drawn in Sec. \ref{sec:conclusion}.

\section{Related Work}
\label{sec:related}
Compared to traditional face hallucination schemes \cite{wang2014comprehensive}, deep learning-based approaches, particularly convolutional neural networks (CNNs),  have proven to achieve state-of-the-art performances in face hallucination \cite{fhgan01,fhgan02_ultra,zhu2016deep,wu2016deep,song-ijcai17-faceSR,bulat2017super,aafh}. For example, a deep learning-based approach to joint face hallucination and recognition was proposed in \cite{wu2016deep}, which involves a SR network and face recognition network. The two networks are jointly optimized iteratively to achieve joint face hallucination and recognition. However, it adopts a relatively shallow CNN to hallucinate face images, resulting in possibly unsatisfactory visual quality of the reconstructed faces. In contrast, \cite{zhu2016deep} proposed a much deeper CNN to generate HR face image. To effectively upscale a LR face without introducing annoying artifacts, the method learns dense correspondence during training and upscale the LR face progressively by a cascading process. During the cascaded iteration, the dense correspondence field is progressively refined with the increased face resolution, while the image resolution is adaptively upsampled as guided by the finer dense correspondence field. To improve the fidelity of a hallucinated HR face, a two-stage method was proposed in \cite{song-ijcai17-faceSR}, that reconstructs facial parts by using a deep CNN, followed by a fine-grained facial structure learner to further refine the reconstructed faces.

Recently, generative adversarial network (GAN) based approaches have been successfully applied to various image processing applications such as image synthesis, image SR, and facial image generator \cite{gan}. A GAN is composed of a generator network and a discriminator network, in which the generator produces image contents based on a learned probability model, whereas the discriminator judges whether the generated contents ares real or fake to decide to accept or reject the contents accordingly. By iterating the adversarial learning process between the generator and the discriminator, the generator will eventually be able to hallucinate photo-realistic image contents that successfully confuse the discriminator.  

For example, the SR GAN (SRGAN) proposed in \cite{gansr} was among the first to infer photo-realistic high-resolution natural images for image SR. In SRGAN, a perceptual loss function consisting of an adversarial loss term and a content loss term was proposed to push the solution to the natural image manifold using a discriminator network that is trained to differentiate between the super-resolved images and original photo-realistic images. This method is, however, not suitable for super-resolving LR face images as explained in \cite{fhgan02_ultra}. To overcome this problem, in \cite{fhgan02_ultra} a pixel-wise $L_2$ regularization term is introduced to the generative model and exploit the feedback of the discriminative network to make the upsampled face images more similar to real ones. Similarly, the method proposed in \cite{fhgan01} utilizes deconvolutional layers to separately super-resolve the local and global parts and uses the discriminator to measure the visual quality of the hallucinated face image. The above-mentioned methods \cite{fhgan01,fhgan02_ultra}, however, cannot guarantee faithful identity preservation of the reconstructed face since they do not provide any identity-preserving guidance to the learning of the discriminator/generator pair. Moreover, they often generate unrealistic reconstructed faces when the resolution of input LR face image is extremely low, as much of facial structure information has been lost. 

 Similarly, \cite{bulat2017super} proposed an end-to-end GAN-based SR scheme which is combined with a face alignment network. The method utilizes heatmap loss to incorporate facial structural information by detecting facial landmarks so as to improve face hallucination results. The deep reinforcement learning method proposed in \cite{aafh} hallucinates HR faces in an iterative reconstruction manner, that employs a recurrent policy network to reconstruct individual HR regions of a face based on previous reconstructions, followed by a local enhancement network to further refine facial details by considering the correlations between different facial parts. 
Nevertheless, the methods proposed in \cite{song-ijcai17-faceSR,bulat2017super,aafh} are only focused on hallucinating visually pleasant HR details without considering whether the added HR details are helpful in recognizing the identity of a face.

\section{Siamese GAN (SiGAN) for Identity-Aware Face Hallucination}
\label{sec:sigan}

\begin{figure*}
	\centering
	\includegraphics[width=0.80\textwidth]{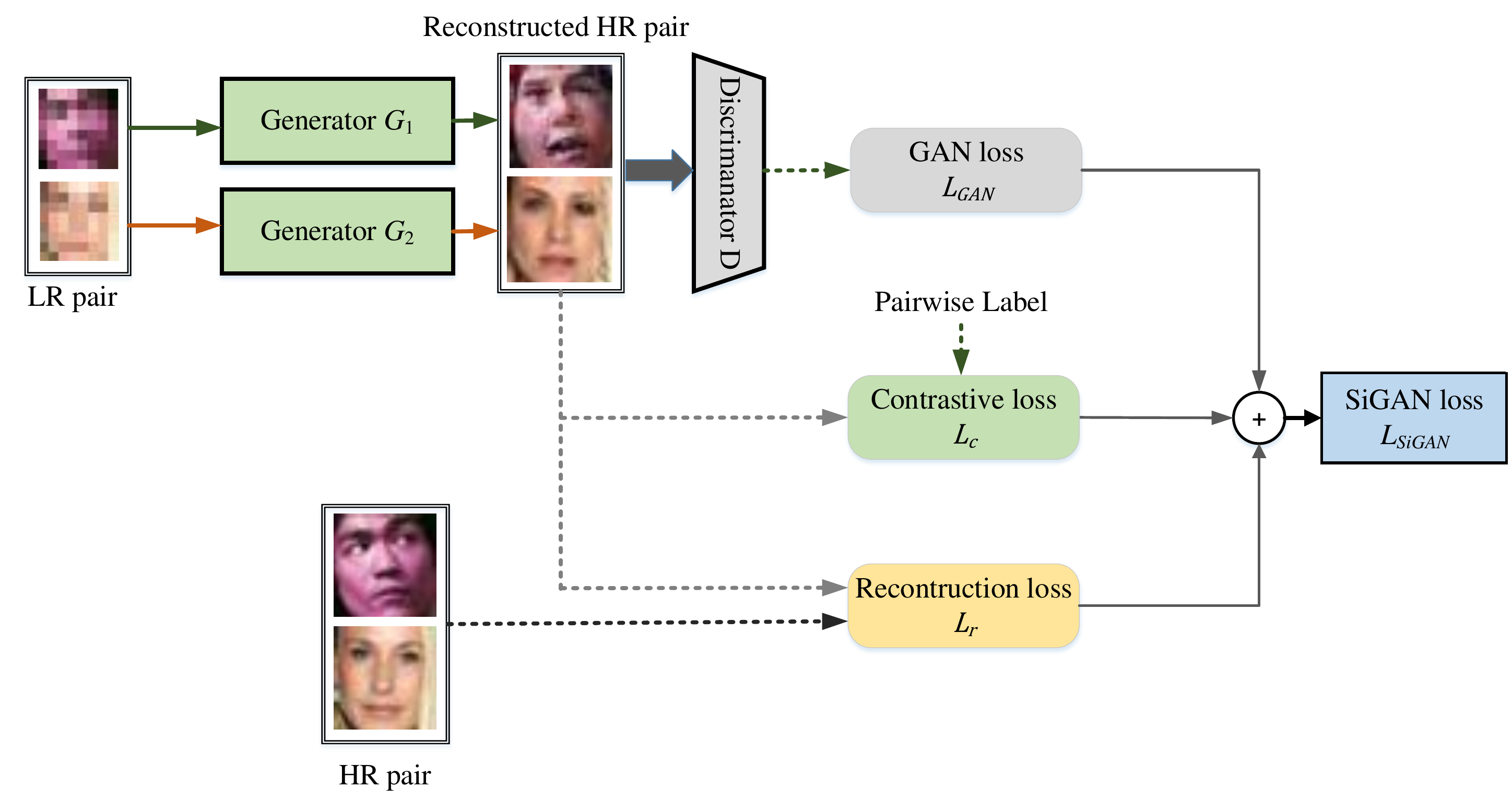}
	\caption{\small Framework of the proposed Siamese GAN (SiGAN) with pairwise identity embedding for face hallucination.}
	\label{fig:frameworkSiGAN}	
\end{figure*}

\subsection{Overview of Proposed SiGAN}

To achieve photo-realistic and identity-preserving reconstruction, the propose SiGAN adopts a pairwise identity learning scheme based on a Siamese network \cite{sia1}, which consists of twin  generators with an identical network model that accept a pair of distinct inputs and are then trained by an energy function at the top. To effectively learn identity-preserving representations, we incorporate in learning the Siamese network an identity-distinguishable contrastive energy function \cite{sia2} which contains dual terms aiming to decrease the energy of same-identity pairs while increasing the energy of different-identity pairs.  Combining the identity-distinguishable contrastive loss with the reconstruction loss terms in SiGAN training can effectively boost the authenticity of reconstructed faces, while achieving good visual fidelity of hallucinated HR faces. 

Note, the adversarial loss incorporated in GAN aims to optimize the following min-max problem \cite{gan} :
\begin{eqnarray}\label{eq:gan}
	\min_{G}\max_{D}V(D,G)=E_{x\sim p_{x}(\mathbf{x})} \left [ \log D(\mathbf{x}) \right ] \\\notag+ E_{z\sim p_{z}(\mathbf{z})}\left [ \log \left (1- D(G(\mathbf{z})) \right ) \right ],
\end{eqnarray}
where $V$ represents the energy function, $D$ and $G$ represent the discriminator and generator, respectively, $G(\mathbf{z})$ is a generated sample from a noise distribution $\mathbf{z}$, and $D(\mathbf{x})$ is the probability of data sample $\mathbf{x}$ being authenticated: $D(\mathbf{x})=1$ indicate that $\mathbf{x}$ is authenticated as a real sample; otherwise $D(\mathbf{x})=0$. 


Fig. \ref{fig:frameworkSiGAN} depicts the framework of SiGAN, which is composed of a pair of two identical spatial-upsampling generators $G_1$ and $G_2$ and a discriminator $D$. In the generator pair, a pair of LR faces is used as prior information to guide HR face generation. Specifically, given a pair of LR faces along with a binary pairwise identity indicator signifying whether the two LR faces belong to a same identity, the contrastive loss term $L_c$ is designed to embed the binary identity label to the generator pair for training. The reconstruction loss $L_r$, defined as the $L_1$ distance between the ground-truth pair and the reconstructed HR face pair, is used to maximize the fidelity of the reconstructed HR face pair. The discriminator then judges whether the generated face is real or fake based on the discriminator loss function $D(\mathbf{x})$ as explained in (\ref{eq:gan}). To train the generator pair and the discriminator, similar to (\ref{eq:gan}), the GAN loss is defined as $L_{GAN}= \log D(\mathbf{x}) + \log \left (1- D(G(\mathbf{z})) \right )$. Consequently, the three loss terms are summed up to obtain the overall loss:  $L_{SiGAN}=L_{GAN}+L_c+L_r$. After training the SiGAN model using an iterative optimization process by minimizing $L_{SiGAN}$ for both the discriminator and the generator pair, we can then use the learned generator to hallucinate HR faces from input LR faces, as elaborated below. 

\begin{figure*}
	\centering
	\includegraphics[width=0.95\textwidth]{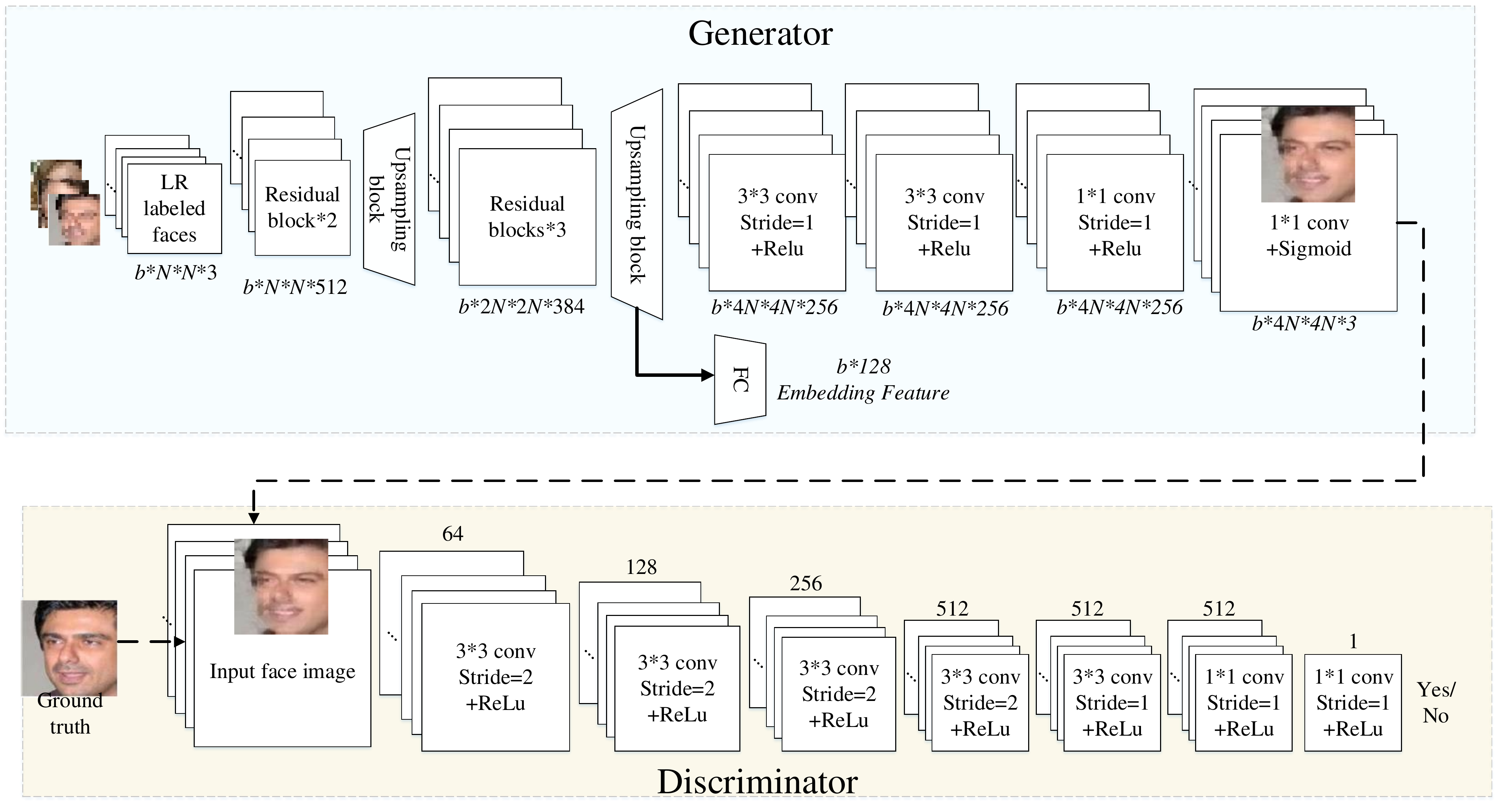}
	\caption{\small Network models of the generator (the upper pipeline) and the discriminator (the lower pipeline) of SiGAN.}
	\label{fig:net_model}
\end{figure*}

\subsection{Network Models}
SiGAN consists of a pair of twin generators, each comprising two/three residual blocks and upsampling blocks, followed by three convolutional layers and a sigmoid function, and a discriminator, which is a fully convolutional network. During training, the generator pair is used to  hallucinate a pair of HR faces from a pair of input LR faces, and the discriminator is used to judge whether the two hallucinated HR faces are real or fake. The generator network and the discriminator network are described below.

\textbf{Generator.}  As shown in the upper pipeline of Fig. \ref{fig:net_model}, the generator is a SR CNN. In the generator, we insert two upsamplers to upscale the input faces by $4\times$. To effectively reconstruct HR faces, we replace the first two layers of the generator of DCGAN \cite{dcgan} with the residual blocks for faster convergence and better training performance. Then, an upsampler is inserted in between the second and the third layers to upscale the input feature maps. The third layer is then followed by three concatenated convolutional layers with a filter size of $3\times3$, and is finally concatenated with a convolutional layer with $1\times1$ kernels. Given an  $N \times N$ face, the size of output face is  $4N\times4N$.

\begin{figure}
	\includegraphics[width=0.5\textwidth]{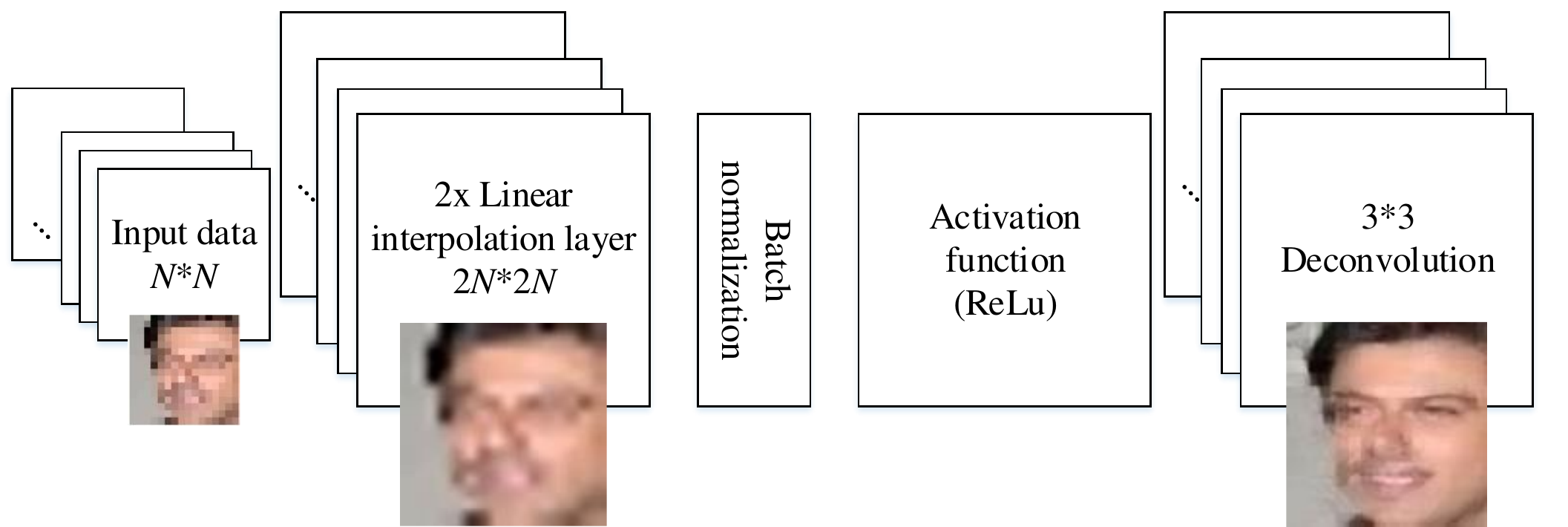}
	\caption{\small Upsampler used in the generator of SiGAN.}
	\label{fig:upsampling}
\end{figure}

\textbf{Upsampler.} Since a CNN usually downscales the input image for extracting feature representations, for upscaling face images, as illustrated in Fig. \ref{fig:upsampling}, we adopt the upsampler proposed in \cite{upsampling} to gradually increase the spatial resolution layer by layer in the CNN. The image size is first linearly interpolated from $N\times N$ to $2N\times 2N$, followed by concatenating a batch normalization and an activation layers. Finally, a deconvolutional layer is used to learn the deconvolution filters to produce a HR face with fine details.

\textbf{Discriminator.} Similar to the discriminator in DCGAN, as illustrated in the lower pipeline in Fig. \ref{fig:net_model}, the discriminator is a fully convolutional network consisting of seven convolutional layers followed by an average polling layer. The output of the discriminator is a normalized value signifying whether the face generated by the generator is true or fake.

\subsection{Training and Optimization}

To learn identity-preserving features while training SiGAN, we incorporate the contrastive loss term into the energy function in (\ref{eq:gan}).  Then, similar to \cite{gansr}, we replace random noise $\mathbf{z}$ in (\ref{eq:gan}) with input LR face $\mathbf{x}^{LR}$. As a result, given ground-truth HR face pair $\mathbf{x}^{HR}_1$ and $\mathbf{x}^{HR}_2$ and the pairwise identity label $\mathbf{y}$, where $y=0$ indicates an impostor pair and $y=1$ indicates a genuine pair, the energy function is defined as
\begin{eqnarray}\label{eq:eng_sigan}
	\min_{G}\max_{D}V(D,G)=E_{D} \left [ \log D(\mathbf{x}^{HR}_1) \right ] \\\notag+ E_{G}\left [ \log \left (1- D(G(\mathbf{x}^{LR}_1)) \right ) \right ]+E_{C}\left [G(\mathbf{x}^{LR}_1), G(\mathbf{x}^{LR}_2)\right],
\end{eqnarray}
where $G(\mathbf{x}^{LR})$ is the generative model used for hallucinating HR faces $\mathbf{x}^{SR}$, and $E_C$ is the contrastive loss defined as
\begin{eqnarray}\label{eq:con_sigan}
	E_C=(1-y)L_I(E_w(\mathbf{x}_1^{SR}, \mathbf{x}_2^{SR})) 
	\\\notag
	+ yL_G(E_w(\mathbf{x}_1^{SR}, \mathbf{x}_2^{SR})).
\end{eqnarray}

Directly computing $E_w(\mathbf{x}_1^{SR}, \mathbf{x}_2^{SR})$ by calculating the $l_1$ norm in the pixel domain (i.e., $E_w=||\mathbf{x}_1^{SR} - \mathbf{x}_2^{SR}||_1^1$), however, usually makes the distance sensitive to the variations in pose, lighting, and expression. Therefore, to better capture the semantic similarity for the generated HR faces, we adopt the  perceptual loss by concatenating a $128$-neuron fully connected layer to the end of the second residual block to generate a $128$-d perceptual feature vector $P(\mathbf{x}^{LR})$ of input LR face $\mathbf{x}^{LR}$. Consequently, we have $E_w=||P(\mathbf{x}_1^{LR}) - P(\mathbf{x}_2^{LR})||_1^1$, $L_I=\frac{1}{2}[\max{(0, m-E_w)}]^2$, $L_G=\frac{1}{2}(E_W)^2$, and $m=0.5$.

Note that, the contrastive loss term not only minimizes the marginal loss $L_I$ between the reconstructed impostor pair $\mathbf{x}^{SR}_1$ and $\mathbf{x}^{SR}_2$, but also minimizes the loss $L_G$ between the super-resolved genuine pair. If the reconstructed HR faces belong to different identities (i.e., $y=0$), minimizing the contrastive loss $E_C$ is equivalent to minimizing $L_I$. By solving (\ref{eq:con_sigan}), we can update the generator toward producing a better identity-preserving reconstruction.

We train SiGAN  by iteratively optimizing the discriminator, generator, and contrastive loss functions using the stochastic gradient descent (SGD) algorithm proposed in \cite{adam}. In each iteration of optimization, we first update the discriminator by ascending its stochastic gradient calculated by 
\begin{eqnarray}\label{eq:sigan-discriminator}
	\triangledown_{\theta_{d}}\frac{1}{b}\sum_{i=1}^{b}\left [ \log D(\mathbf{x}^{HR}_1) \right ] + \left [ \log \left (1- D(G(\mathbf{x}^{LR}_1)) \right ) \right ].
\end{eqnarray}
Then, we update the generator pair by descending its gradient calculated by
\begin{eqnarray}\label{eq:sigan-generator}
	\triangledown_{\theta_{g}}\frac{1}{b}\sum_{i=1}^{b} \log \left (1- D(G(\mathbf{x}^{LR}_1)) \right ).
\end{eqnarray}
Finally, we fix the updated results of the generator pair and discriminator, and update the generator pair based on the contrastive loss function by descending its gradient:
\begin{eqnarray}\label{eq:sigan-contrastive}
	\triangledown_{\theta_{c}}\frac{1}{b}\sum_{i=1}^{b}(1-y)L_I(E_w(P(\mathbf{x}_1^{LR}), P(\mathbf{x}_2^{SR}))) 
	\\\notag
	+ yL_G(E_w(P(\mathbf{x}_1^{LR}), P(\mathbf{x}_2^{LR}))).
\end{eqnarray}
Taking several training epochs of the proposed SiGAN using SGD, we can learn the model of the generator pair that can hallucinate photo-realistic and identity-preserving HR faces. 

\section{Face Hallucination GANs with Direct Identity Embedding}
\label{sec:fhgan}

\begin{figure*}
	\begin{center}	\includegraphics[width=0.9\textwidth]{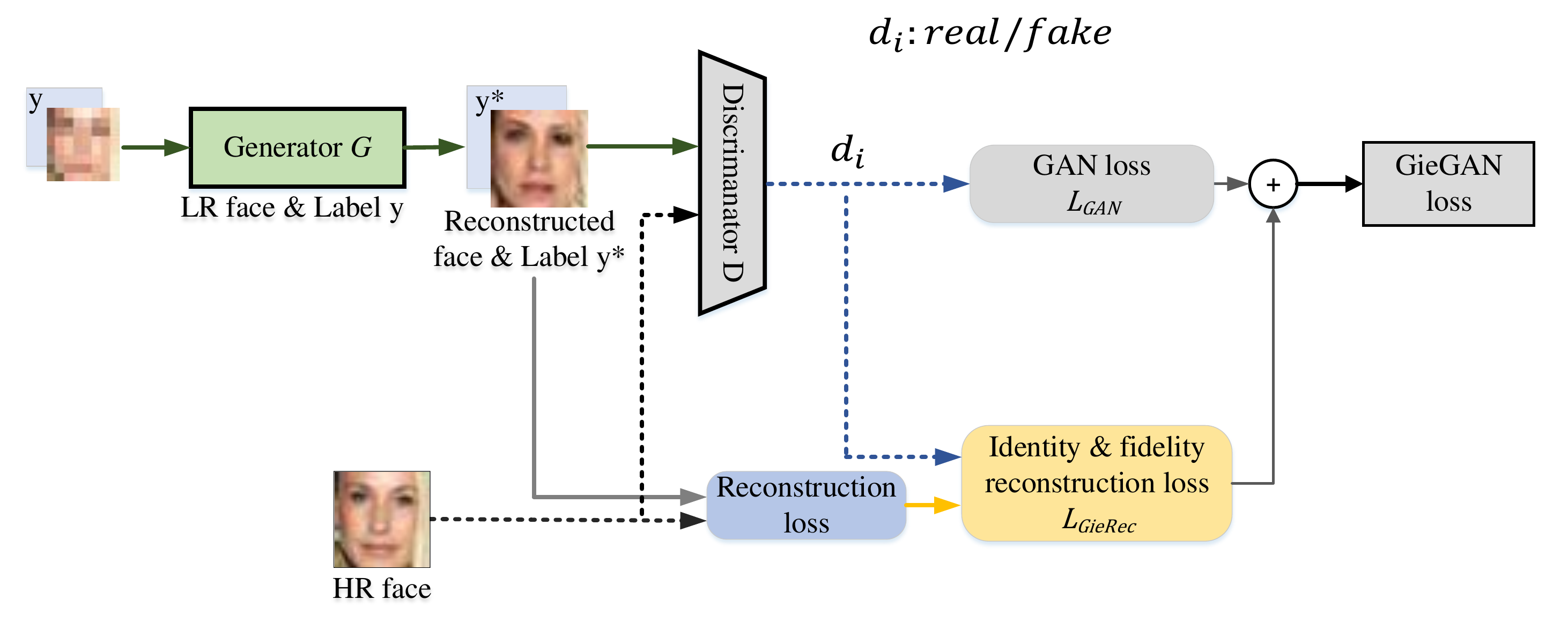}
		
	\text{(a)}\\
	\vspace{0.1cm}
	\includegraphics[width=0.9\textwidth]{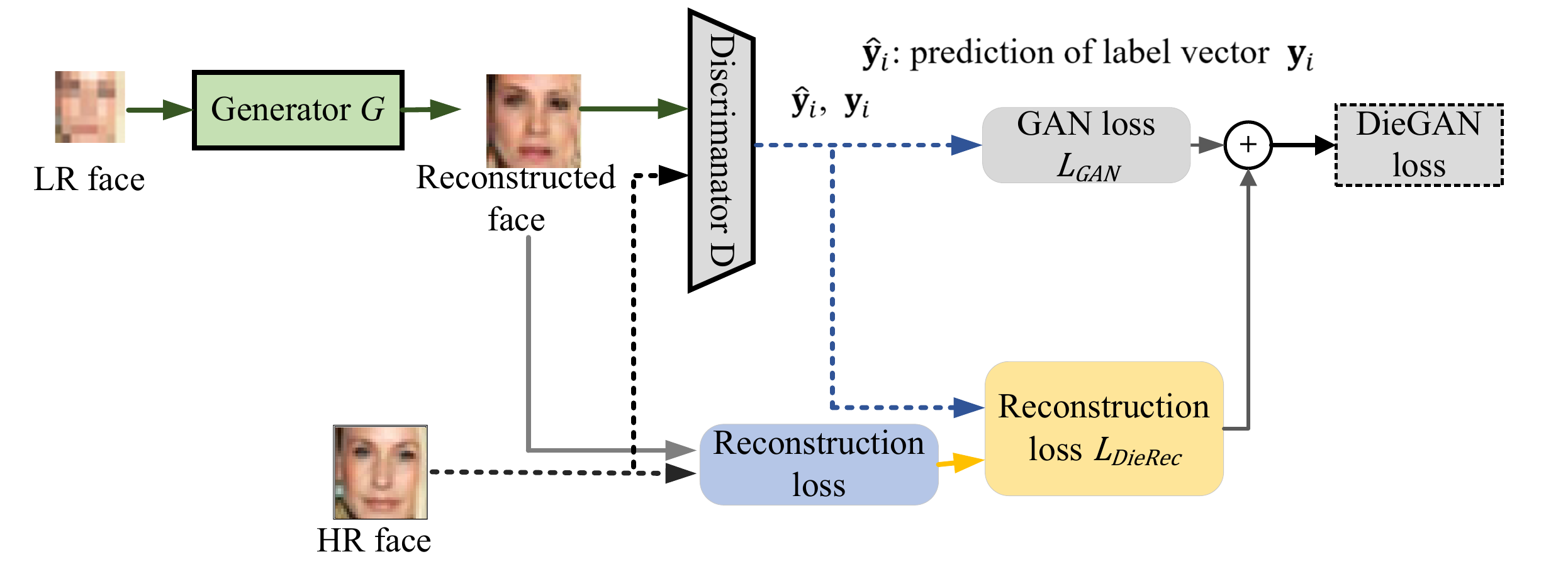}
	
	\text{(b)}\\
	\vspace{0.1cm}
	\end{center}
	\caption{\small Block diagrams of (a) the Generator identity embedding GAN (GieGAN) and (b) the Discriminator identity embedding GAN (DieGAN).}
	\label{fig:fhgan}	
\end{figure*}


In practice, there are multiple ways of embedding identity information in face hallucination GANs. Besides the proposed SiGAN, for the sake of comparison, we also design two variants of face hallucination GANs with direct identity embedding: the Generator identity embedding GAN (GieGAN) and the Discriminator identity embedding GAN (DieGAN) as depicted in Fig. \ref{fig:fhgan}, respectively. Both GieGAN and DieGAN are built on top of DCGAN by additionally incorporating label information and  reconstruction loss in network training to achieve photo-realistic and identity-preserving hallucination. The major difference between SiGAN and GieGAN/DieGAN is that SiGAN learns identity-preserving representations through "weak" identity embedding in a pairwise learning manner (i.e., only needs a simple label indicating whether a pair of training faces belong to a same person, instead of their exact identity labels), whereas GieGAN and DieGAN have to learn from exact identity labels in GAN training, making their labeling cost much higher compared to SiGAN.   

\subsection{Generator identity embedding GAN (GieGAN)}
\label{GieGAN}
As depicted in Fig. \ref{fig:fhgan}(a), GieGAN is composed of a spatial-upsampling generator $G$ and a discriminator $D$. In the generator, the input LR face is first appended with an additional identity channel that contains a normalized identity label value. To match the dimension of the RGB channels, the scalar identity label is expended to a vector with the same dimension of the input LR face by replicating its value to all the entries of the vector. Similarly, the output of the generator also consists of four channels: the RGB channels  and the identity label of the reconstructed HR face.

To train GieGAN, we modify the energy function in (\ref{eq:gan}) by replacing the distribution $G(\mathbf{z})$ in (\ref{eq:gan}) with $G(\mathbf{z}|\mathbf{y})$ to condition the generative model on some external information. Then, similar to \cite{gansr}, we replace random noise $\mathbf{z}$ in (\ref{eq:gan}) with input LR face $\mathbf{x}^{LR}$. As a result, given the ground-truth HR face $\mathbf{x}^{HR}$ and the face's identity label $\mathbf{y}$, the energy function is expressed as
\begin{eqnarray}\label{eq:giegan}
	\min_{G}\max_{D}V(D,G)=E_{x\sim p_{x}(\mathbf{x})} \left [ \log D(\mathbf{x}^{HR}) \right ] \\\notag+ E_{z\sim p_{z}(\mathbf{z}|\mathbf{y})}\left [ \log \left (1- D(G(\mathbf{x}^{LR}|\mathbf{y})) \right ) \right ],
\end{eqnarray}
where $G(\mathbf{x}^{LR}|\mathbf{y})$ is the generative model used for hallucinating HR faces $\mathbf{x}^{SR}_y$.

To ensure photo-realistic and identity-preserving and photo-realistic hallucination, given the output values of the discriminator for a training batch with a batch size of $b$, the overall loss function for training the generator and discriminator of GieGAN can be defined as the sum of an realism loss term, a reconstruction loss term, and a GAN loss term as follows: 
\begin{eqnarray}\label{eq:gierec_loss}
L_{GieLoss}=\gamma L_{Real}+ \beta L_{Rec}+(1-\gamma - \beta)L_{GAN},
\end{eqnarray}
where $\gamma$, $\beta$, and $(1-\gamma - \beta)$ represent the weights for the realism loss, reconstruction loss, and GAN loss, respectively. 

The realism loss  $L_{Auth}$ measures, as judged by the discriminator, how realistic a hallucinated HR face is. It is defined as the cross-entropy between the binary judgments  (real or fake) of the discriminator and their ideal outcomes (always real) for a batch of training faces:
\begin{eqnarray}\label{eq:gieauthen_loss}
	L_{Real}=-\frac{1}{b}\sum_{i=1}^b d^*_i \log(d_i)-(1-d^*_i) \log(1-d_i)\\\notag=-\frac{1}{b}\sum_{i=1}^b  \log(d_i),
\end{eqnarray}
where $d_i$ is the output value of the discriminator for the $i$-th training face in a training batch and $d^*_i$ is 1 (ideally judged as a real face). 

The reconstruction loss of the generator is defined as the $L1$ norm of the difference between the appended ground-truth and its hallucinated version:
\begin{eqnarray}\label{eq:gieid_loss}
L_{Rec}=\frac{1}{b}\sum_{i=1}^b \left \|  \mathbf{x}^{HR}_{y.i}-\mathbf{x}^{SR}_{y,i} \right \|_{1},
\end{eqnarray}
where $\mathbf{x}^{HR}_{y.i}$ and $\mathbf{x}^{SR}_{y.i}$ respectively denote the expanded HR ground-truth and its hallucinated version of the $i$th training face. Since an an additional identity label channel is appended, the term measures the reconstruction loss in both fidelity and identity. As a result, identity-preserving representations are learned through this loss term.

	Similar to (\ref{eq:gan}), the GAN loss is defined as $L_{GAN}= \log D(\mathbf{x}^{HR}) + \log \left (1- D(G(\mathbf{x}^{LR}|\mathbf{y})) \right )$. By minimizing the overall loss function of GieGAN, we not only keep the high fidelity of reconstructed HR faces but also restore their identity information based on the two facts: $1)$ The side information (i.e., the identity label) constrains the solution space of the generator to maximize the relevance of hallucinated faces to their corresponding identity, and $2)$ the reconstruction loss term maximizes the fidelity of the HR face hallucinated by the generator. As a result, the hallucinated HR faces are both photo-realistic and identity-preserving.

Since, unlike the training phase, the identity label is usually unavailable with the input LR face, we propose an approach to hallucinate a HR face without label information. With GieGAN, the ideal goal is to reconstruct a HR face with the correct identity label that the discriminator cannot judge its authenticity. Otherwise, if the training LR face is associated with wrong label information, the discriminator will reject the hallucinated face. The confidence score of unlabeled input LR face $\mathbf{x}^{LR}$ is calculated by the discriminator as $A_y=D(\mathbf{x}^{LR}|y)$. The larger $A_y$  is, the more realistic $\mathbf{x}^{SR}$ will be.  We search all possible identity labels to find the identity label with the highest confidence score as follows:
\begin{eqnarray}
	\arg\max_{i}A_{y}(i)=D(\mathbf{x}^{LR}|\mathbf{y_{i}})\  \forall{i}\subseteqq{\mathbf{I}}
	\label{eq:test}
\end{eqnarray}
where  $\mathbf{I}=\{0,1,...,C\}$ denotes the set of possible identity labels and $C$ is the number of identity classes.
Consequently, the most possible label is identified and the reconstructed face will be the best one.

\subsection{Discriminator identity embedding GAN (DieGAN)} 
\label{DieGAN}
Different from GieGAN, in DieGAN, the identity information is embedded in the discriminator rather than the generator. In this way, the HR faces can be hallucinated without the need of searching over all possible identity labels. As depicted in Fig. \ref{fig:fhgan}(b), the generator of DieGAN is similar to that of GieGAN but the label channel is removed from the input LR faces. The discriminator of DieGAN not only distinguishes whether a face is real or fake but also predicts its identity label $y$. We modify the discriminator of SiGAN to handle multi-class prediction by expanding the number of channels in the last convolutional layer to $C+1$, where $C$ represents the number of identity classes in the training data, and the additional class is used to indicate fake HR faces. In our implementation, the discriminator is a fully convolutional network consisting of 10 convolutional layers followed by an average polling layer. Then, similar to (\ref{eq:gierec_loss}), we define the reconstruction loss of DieGAN as the sum of an identity loss term and a fidelity loss term as follows:
\begin{eqnarray}\label{eq:DieG_loss}
	L_{DieRec}=\frac{1}{b}\sum_{i=1}^b \{(1-\gamma)\left [-\mathbf{y}_i \cdot \log(\widehat{\bf{y}}_i)\right ]+\gamma \left \|  \mathbf{x}_i^{SR}-\mathbf{x}_i^{HR} \right \|_{1}\},
\end{eqnarray}
where identity label vector $\mathbf{y}_i$ associated with the $i$-th training face in a training batch is a ($C+1$)-dimentional one-hot vector, $\widehat{\bf{y}}_i$ stands for the prediction of $\mathbf{y}_i$ by the discriminator of DieGAN. Finally, we apply standard Adam SGD \cite{adam} to iteratively minimize the overall loss $L_{DieGAN} =L_{GAN}+ L_{DieRec}$.

\section{Experimental Results}
\label{sec:experiments}
For performance evaluation, we compare our identity-embedding methods (SiGAN, GieGAN, and DieGAN) with several existing methods including bicubic interpolation, ultra-resolution by discriminative generative networks (UR-DGN) \cite{fhgan02_ultra}, deep facial component generation method (DFCG) \cite{song-ijcai17-faceSR}, DCGAN \cite{dcgan}, and pixel recurrent super-resolution (PRSR) \cite{pixelrecurrent}. Since there is still no widely-accepted objective quality metric for face hallucination currently, besides subjective evaluation, we further perform face recognition and verification on reconstructed HR faces using state-of-the-art OpenFaces engine  \cite{openface}, and use the face recognition/versification rate as an objective quality metric to evaluate whether the reconstructed HR details are useful for identity recognition. The compared methods are all trained and tested on a publicly available face dataset CASIA-WebFace \cite{casia} or simply CASIA. Besides the CASIA dataset, we also do performance evaluation against two faces-in-the-wild datasets: the Labeled Faces in the Wild (LFW) \cite{lfw} and CelebA \cite{celebA}. All face images are cropped to the size of $128\times128$ without any further preprocessing. The size of input LR face images is downscaled to $8\times8$ and $16\times16$ and then superresolved to $32\times32$ and $64\times64$, respectively, by various face hallucination schemes.

\subsection{Subjective Visual Quality Evaluation}

\begin{figure*}
	\centering
	\includegraphics[width=0.92\textwidth]{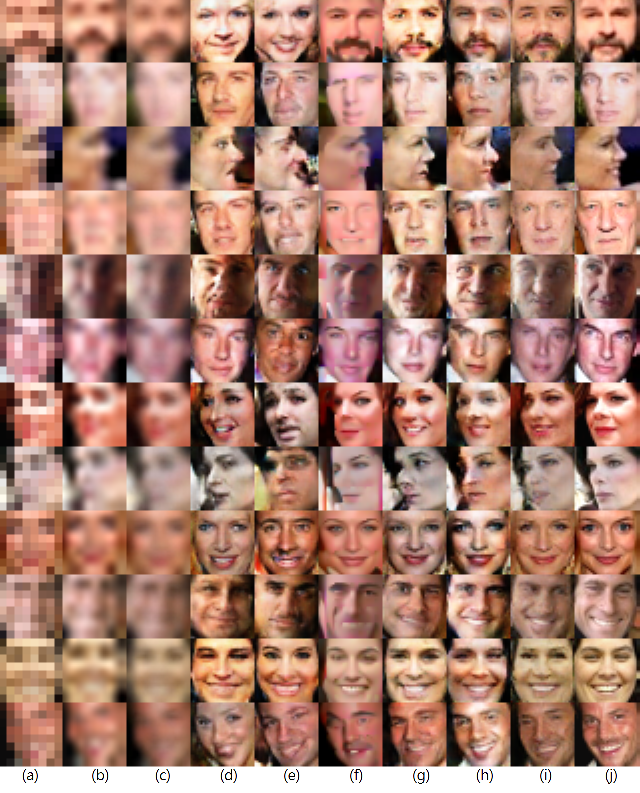}
	\caption{\small Subjective visual quality comparison of various face hallucination methods for 12 identities selected from CASIA \cite{casia}: (a) The LR face images ($8 \times 8$).  (b)--(i) are the reconstructed $32 \times 32$ HR faces using (b) bicubic interpolation, (c) DFCG \cite{song-ijcai17-faceSR}, (d) DCGAN \cite{dcgan}, (e) UR-DGN \cite{fhgan02_ultra}, (f) PRSR \cite{pixelrecurrent}, (g) GieGAN , (h) DieGAN, (i) SiGAN, and (j) the ground-truths ($32 \times 32$.)}
	\label{fig:casia8x8}	
\end{figure*}
\begin{figure*}
	\centering
	\includegraphics[width=0.92\textwidth]{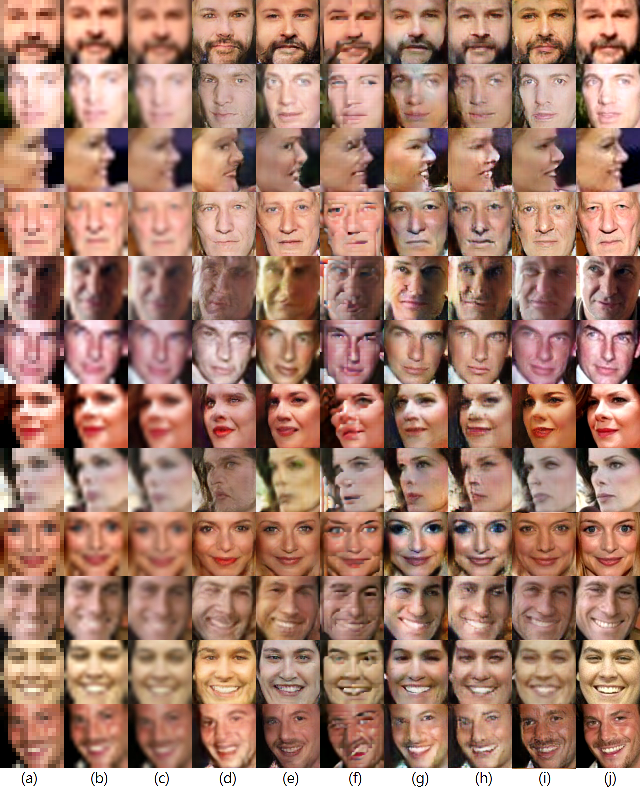}
	\caption{\small Subjective visual quality comparison of various face hallucination methods for 12 identities selected from CASIA \cite{casia}: (a) The LR face images ($16 \times 16$).  (b)--(h) are the reconstructed $64 \times 64$ HR faces using (b) bicubic interpolation, (c) DFCG \cite{song-ijcai17-faceSR}, (d) DCGAN \cite{dcgan}, (e) UR-DGN \cite{fhgan02_ultra}, (f) GieGAN, (g) DieGAN, (h) SiGAN (proposed), and (i) the ground-truths ($64 \times 64$.)}
	\label{fig:casia16x16}	
\end{figure*}

\textbf{CASIA Dataset.} The CASIA  dataset \cite{casia} contains $494,414$ face images with various illuminations and poses captured from $10,575$ subjects. In each trial, we randomly select $491,131$ out of  the $494,414$ face images for training and use the remaining $3,283$ images for testing. Fig. \ref{fig:casia8x8} illustrates the face hallucination results for $12$ test faces upscaled from  $8\times8$ to $32\times32$. In Fig. \ref{fig:casia8x8}, since the resolution of the LR faces is only $8\times8$, most of detailed facial information is missing. As a result, we can observe that the HR faces reconstructed by DFCG \cite{song-ijcai17-faceSR} show over-smooth results because the LR observations lack enough information for correctly estimating the initial facial parts, making the refiner in DFCG fail to well hallucinate the HR details of facial parts.  In contrast, although the DCGAN-based approach can hallucinate photo-realistic HR faces, the reconstructed HR faces are usually significantly dissimilar to their corresponding identities, as neither reconstruction loss nor identity information is considered in DCGAN. In contrast, UR-DGN \cite{fhgan02_ultra} takes into account reconstruction loss in the CNN training to improve the fidelity of reconstructed HR faces, which, however, still often reconstructs HR faces with significantly dissimilar facial parts compared with their ground-truths due to the lack of identity information. Although PRSR \cite{pixelrecurrent} can produce fine and smooth details, it may generate severe artifacts if the initial HR face is not well inferred, which often causes serious error propagation in the succeeding step-by-step refinement.  Besides, the lack of identity information in PRSR will also make the reconstructed HR faces unrecognizable in identity. Since SiGAN takes into account both the reconstruction loss and label information to overcome the above problems, besides successfully hallucinating the fine details, the reconstructed HR facial parts more faithfully resemble their corresponding ground-truths. The generator-based identity embedding scheme, GieGAN, though also achieving photo-realistic visual quality, reconstructs less faithful facial parts compared to SiGAN, whereas the discriminator-based scheme, DieGAN, produces more severe artifacts on the reconstructed HR faces compared to SiGAN and GieGAN.  Fig. \ref{fig:casia16x16} illustrates the HR faces hallucinated from $16\times16$ to $64\times64$ for the same test faces in Fig. \ref{fig:casia8x8}. Again, the results show that SiGAN outperforms the other schemes in both visual fidelity and authenticity of the reconstructed HR faces.

\begin{figure*}
	\centering
	\includegraphics[width=0.92\textwidth]{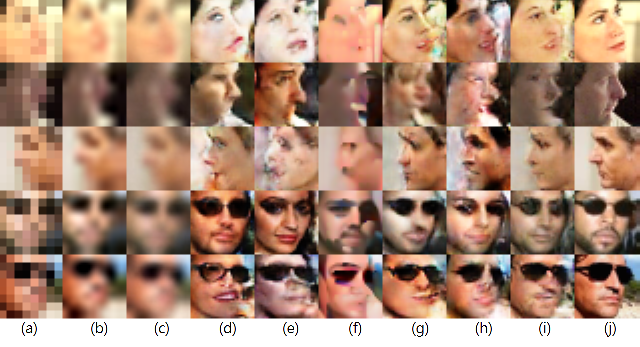}
	\caption{\small Subjective visual quality comparison for five faces with unknown identities selected from LFW \cite{lfw} and CelebA \cite{celebA}: (a) The LR face images ($8 \times 8$).  (b)--(i) are the reconstructed $32 \times 32$ HR faces using (b) bicubic interpolation, (c) DFCG \cite{song-ijcai17-faceSR}, (d) DCGAN \cite{dcgan}, (e) UR-DGN \cite{fhgan02_ultra}, (f) PRSR \cite{pixelrecurrent}, (g) GieGAN , (h) DieGAN, (i) SiGAN, and (j) the ground-truths ($32 \times 32$).}
	\label{fig:lfw8x8}	
\end{figure*}

\begin{figure*}
	\centering
	\includegraphics[width=0.92\textwidth]{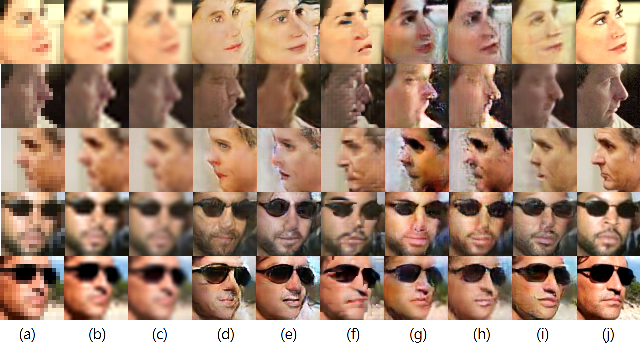}
	\caption{\small Subjective visual quality comparison for five faces with unknown identities selected from LFW \cite{lfw} and CelebA \cite{celebA}: (a) The LR face images ($16 \times 16$).  (b)--(i) are the reconstructed $64 \times 64$ HR faces using (b) bicubic interpolation, (c) DFCG \cite{song-ijcai17-faceSR}, (d) DCGAN \cite{dcgan}, (e) UR-DGN \cite{fhgan02_ultra}, (f) PRSR \cite{pixelrecurrent}, (g) GieGAN, (h) DieGAN, (i) SiGAN, and (j) the ground-truths ($64 \times 64$.)}
	\label{fig:lfw16x16}	
\end{figure*}

\textbf{Faces in The Wild Datasets.} Since in many applications the input LR faces often belong to unknown identities, we also evaluate the performances of hallucination methods on faces whose identities are not included in the training set to verify if these methods can be generalized to input faces with unknown identities. In the experiment, we randomly sample face images from two face-in-the-wild datasets, LFW \cite{lfw} and CelebA \cite{celebA}, as test images to evaluate the generality of the compared methods which are all trained on the CASIA dataset. Fig. \ref{fig:lfw8x8} illustrates the  $8\times8$ to $32\times32$ face hallucination results of five difficult test faces (e.g., faces wearing glasses and non-frontal faces) selected from LFW \cite{lfw} and CelebA \cite{celebA}. We can observe that all methods produce a few artifacts on the HR faces, because the numbers of training samples for such types of faces are very limited, making the generator difficult to train well for the face structures. For example, the fifth test face not only wears glasses but also involves some background information. In this case, all methods fail to hallucinate correct HR facial parts. Nevertheless, compared to the other methods, SiGAN still achieves significantly better visual qualities.   Fig. \ref{fig:lfw16x16} shows the HR faces hallucinated from $16\times16$ to $64\times64$ for the same identities in Fig. \ref{fig:lfw8x8}. SiGAN achieves the best performance as well.

\subsection{Objective Quality Evaluation Based on Face Recognition/Verification}
To evaluate the degree of authenticity of reconstructed HR faces compared to their ground-truth identity, we use a state-of-the-art CNN-based face recognition engine, OpenFaces \cite{openface}, to evaluate the face recognition rate and verification rate for HR faces reconstructed by various face hallucination methods. We adopt two objective evaluation approaches. First, we employ OpenFaces \cite{openface} trained from training HR faces of CASIA to recognize the identities of the reconstructed HR faces and calculate the identity recognition rate. Second, following the standard face verification methodology described in  \cite{openface}, based on pair matching, we evaluate the accuracy of reconstructed HR faces being verified by OpenFaces as the same identity with their corresponding ground-truth face. Both these two strategies are used to evaluate the objective  performances of various face hallucination methods against CASIA \cite{casia} and LFW \cite{lfw}. Since CelebA does not provide identity labels, it is not used in the objective evaluation.

\subsubsection{Face recognition performance comparison}
For the experiments on CASIA, we randomly sample $144,942$ images belonging to $671$ identities to train OpenFaces. We then sample $2,000$ face images from the remaining images as the test dataset to evaluate the face recognition performance. Since the number of face images of some identities in CASIA is small, we only choose those identities with more than 120  face images in the dataset, as suggested in \cite{robustFR}. For the experiment on LFW, we first randomly sample $11,000$ face image belonging to $680$ identities as the training set, and sample $2,000$ face images from the remaining as the test dataset. To train OpenFaces, all face images are resized to $96\times 96$, as suggested in \cite{openface}. Similarly, in the testing stage, all hallucinated HR faces and LR faces are resized to $96\times96$.

We first evaluate the face recognition performances on hallucinated HR faces associated with identities that are included in the training set. Table \ref{tab:recognition_casia}(a) compares the top-1, top-5, and top-10 face recognition rates for $32 \times 32$ HR faces upscaled from $8 \times 8$ LR faces using various methods.  The result shows that, as evaluated by OpenFaces, the average recognition rates for the HR faces reconstructed by SiGAN and GieGAN are significantly higher than those achieved by the other methods. Besides, DieGAN performs slightly worse than SiGAN and GieGAN do, but still significantly outperforms the remaining methods, because many identities (say, $10,575$ in CASIA) need to be learned in the discriminator of DieGAN, thereby making it relatively difficult to train. Among the existing methods, compared to bicubic interpolation, UR-DGN \cite{fhgan02_ultra} achieves slightly lower face recognition rate, whereas DFCG \cite{song-ijcai17-faceSR}, DCGAN \cite{dcgan}, and PRSR \cite{pixelrecurrent} all significantly degrade face recognition performance, meaning that the HR details reconstructed by these methods are not useful and even usually incorrect for identity recognition.  Table \ref{tab:recognition_casia}(b) compares the average face recognition rates for $64 \times 64$ HR faces upscaled from $16 \times 16$ LR faces using various methods. Similarly, SiGAN and GieGAN achieve the best average recognition rates, and the face recognition rate with DieGAN is slightly lower than that with SiGAN and GieGAN, but higher than the remaining.

Since in many applications the input LR faces usually belong to unknown identities, Table \ref{tab:recognition_lfw} compares the performances of various hallucination methods on faces randomly sampled from LFW whose identities are not included in the training set of CASIA used for training OpenFaces to verify the generality of these methods to faces belonging to unknown identities. Again, SiGAN  achieves the best average recognition rates, showing that even for faces with unknown identities, SiGAN can still effectively enhance identity-preserving facial details.

\begin{table}
	\caption{\small Comparison of face recognition rates evaluated by OpenFaces \cite{openface} for HR faces reconstructed by various face hallucination methods on CASIA \cite{casia} by upscaling: (a) from $8\times 8$ to $32\times 32$; (b) from $16\times 16$ to $64\times 64$}
	\label{tab:recognition_casia}
	\begin{center}
		\text{(a)}\\
		\vspace{0.1cm}
		\begin{tabular}{l|r|r|r}
			\hline
			Method & Top-1 & Top-5 & Top-10 \\
			\hline\hline
			HR ($32 \times 32$)			& 30.4\% & 51.2\% & 59.6\% \\
			LR ($8 \times 8$ )			&  10.7\% & 19.5\% & 33.1\% \\
			Bicubic 					&  10.8\% & 20.1\% & 34.4\% \\
			DFCG \cite{song-ijcai17-faceSR} 		&  9.3\% &  17.7\% & 21.4\% \\
			UR-DGN \cite{fhgan02_ultra} &  9.9\% &  18.6\% & 22.7\% \\
			DCGAN \cite{dcgan} 			&  4.6\% &  10.9\% & 16.8\% \\
			PRSR \cite{pixelrecurrent} 	&  10.8\% &  18.8\% & 24.4\% \\
			GieGAN  						&  14.3\% & 26.6\% & 39.6\% \\
			DieGAN 						&  12.4\% & 25.1\% & 37.5\% \\
			SiGAN (proposed)			&  15.8\% & 27.5\% & 40.4\% \\
			\hline
		\end{tabular}
	\end{center}

	\begin{center}
		\text{(b)}\\
		\vspace{0.1cm}
	\begin{tabular}{l|r|r|r}
		\hline
		Method & Top-1 & Top-5 & Top-10 \\
		\hline\hline
		HR ($64 \times 64$)         & 36.8\% & 55.9\% & 63.8\% \\
		LR ($16 \times 16$)         & 12.4\% & 27.4\% & 37.1\% \\
		Bicubic                     & 11.6\% & 27.5\% & 37.6\% \\
		DFCG \cite{song-ijcai17-faceSR}      &  9.6\% &  23.7\% & 34.8\% \\
		UR-DGN \cite{fhgan02_ultra} & 12.2\% & 29.0\% & 38.7\% \\
		DCGAN \cite{dcgan}          & 9.3\%  & 24.9\% & 33.9\% \\
		PRSR \cite{pixelrecurrent}  & 13.3\% & 29.7\% & 40.1\% \\
		GieGAN       	            & 17.0\% & 36.3\% & 46.4\% \\
		DieGAN       	            & 13.3\% & 31.0\% & 40.7\% \\
		SiGAN (proposed)            & 17.9\% & 32.9\% & 48.1\% \\
		\hline
	\end{tabular}
\end{center}
\end{table}

\begin{table}
	\caption{\small Comparison of face recognition rates evaluated by OpenFacses \cite{openface} for HR faces reconstructed by various face hallucination methods on LFW \cite{lfw} by upscaling: (a) from $8\times 8$ to $32\times 32$; (b) from $16\times 16$ to $64\times 64$}
	\label{tab:recognition_lfw}
	\begin{center}
		\text{(a)}\\
		\vspace{0.1cm}
		\begin{tabular}{l|r|r|r}
			\hline
			Method & Top-1 & Top-5 & Top-10 \\
			\hline\hline
			HR ($32 \times 32$)			&  32.2\% & 50.8\% & 56.7\% \\
			LR ($8 \times 8$ )			&  9.3\% & 17.4\% & 30.9\% \\
			Bicubic 					&  9.6\% & 17.7\% & 30.4\% \\
			DFCG \cite{song-ijcai17-faceSR} 		&  9.3\% &  16.9\% & 27.5\% \\
			UR-DGN \cite{fhgan02_ultra} &  7.9\% &  16.8\% & 20.1\% \\
			DCGAN \cite{dcgan} 			&  4.7\% &  9.9\% &  14.6\% \\
			PRSR \cite{pixelrecurrent} 	&  10.3\% & 19.8\% & 26.1\% \\
			GieGAN  					&  13.9\% & 24.1\% & 37.7\% \\
			DieGAN 						&  13.8\% & 24.6\% & 36.9\% \\
			SiGAN (proposed)			&  14.5\% & 26.7\% & 39.2\% \\
			\hline
		\end{tabular}
	\end{center}

	\begin{center}
		\text{(b)}\\
		\vspace{0.1cm}
	\begin{tabular}{l|r|r|r}
		\hline
		Method & Top-1 & Top-5 & Top-10 \\
		\hline\hline
		HR ($64 \times 64$)         & 35.4\% & 51.4\% & 60.1\% \\
		LR ($16 \times 16$)         & 14.8\% & 26.6\% & 35.3\% \\
		Bicubic                     & 15.0\% & 26.4\% & 35.6\% \\
		DFCG \cite{song-ijcai17-faceSR}      &  13.2\% &  25.4\% & 34.7\% \\
		UR-DGN \cite{fhgan02_ultra} & 15.9\% & 30.2\% & 39.4\% \\
		DCGAN \cite{dcgan}          & 11.6\%  & 24.3\% & 32.6\% \\
		PRSR \cite{pixelrecurrent}  & 18.3\% & 32.6\% & 45.5\% \\
		GieGAN  		               & 20.0\% & 38.4\% & 49.4\% \\
		DieGAN   	                & 19.8\% & 38.4\% & 48.6\% \\
		SiGAN (proposed)            & 21.5\% & 40.5\% & 50.2\% \\
		\hline
	\end{tabular}
\end{center}
\end{table}

\subsubsection{Face verification performance comparison}
In this experiment, we first randomly sample $500,000$ and $200,000$ face pairs from CASIA and LFW, respectively, as the training sets for training the OpenFaces recognition engine with the settings specified in \cite{openface}. We then randomly sample $6,000$ faces from the remaining data samples of CASIA and LFW, respectively, as the test set to evaluate the face verification performance.  

We first evaluate the area under curve (AUC) \cite{lfw} of the trained face verification system for the hallucinated HR faces associated with identities that are included in the training set. Table \ref{tab:casia_result1v} compares the AUCs for $32 \times 32$ and $64\times64$ HR faces respectively reconstructed from $8 \times 8$ and $16\times16$ LR faces using various face hallucination methods. The result shows that, as evaluated by the OpenFaces engine \cite{openface}, the AUC for the HR faces reconstructed by SiGAN is significantly higher than those achieved by the other methods, meaning that SiGAN achieves a significantly higher degree of authenticity of reconstructed HR faces to their ground-truth identity.  Table \ref{tab:lfw_result1v} compares the AUCs of various face hallucination methods on LFW. Again, SiGAN achieves the best AUC performance.

\begin{table}
	\caption{\small Performance comparison evaluated by OpenFaces \cite{openface} for various face hallucination methods on CASIA \cite{casia}}
	\label{tab:casia_result1v}
	\begin{center}
		\begin{tabular}{l|c|c}
			\hline
			
			Methods & $8 \times 8$ to $32 \times 32$ & $16 \times16$ to $64 \times 64$\\
			\hline\hline
			HR 							&  83.3\% & 92.7\%  \\
			LR 							&  64.1\% & 64.3\% \\
			Bicubic 					&  64.8\% & 63.7\%  \\
			DFCG \cite{song-ijcai17-faceSR} 		&  63.7\% & 64.0\%  \\
			UR-DGN \cite{fhgan02_ultra} &  64.5\% & 67.7\%  \\
			DCGAN \cite{dcgan} 			&  60.9\% & 60.8\%  \\
			PRSR \cite{pixelrecurrent} 	&  70.0\% & 71.1\% \\
			GeGAN  						&  76.6\% & 78.4\% \\
			DeGAN 						&  77.9\% & 78.2\% \\
			SiGAN (proposed)			&  81.2\% & 82.8\% \\
			\hline
		\end{tabular}
	\end{center}
\end{table}

\begin{table}
	\caption{\small Performance comparison evaluated by OpenFaces \cite{openface} for various face hallucination methods on LFW \cite{lfw}}
	\label{tab:lfw_result1v}
	\begin{center}
		\begin{tabular}{l|c|c}
			\hline
			
			Methods & $8 \times 8$ to $32 \times 32$ & $16 \times16$ to $64 \times 64$\\
			\hline\hline
			HR 							&  97.6\% & 98.8\%  \\
			LR 							&  70.7\% & 75.4\% \\
			Bicubic 					&  70.8\% & 75.7\%  \\
			DFCG \cite{song-ijcai17-faceSR} 		&  68.6\% & 73.9\%  \\
			UR-DGN \cite{fhgan02_ultra} &  67.7\% & 72.8\%  \\
			DCGAN \cite{dcgan} 			&  64.9\% & 74.8\%  \\
			PRSR \cite{pixelrecurrent} 	&  69.6\% & 76.9\% \\
			GieGAN  					&  77.3\% & 78.6\% \\
			DieGAN 						&  76.1\% & 77.7\% \\
			SiGAN (proposed)			&  82.9\% & 83.4\% \\
			\hline
		\end{tabular}
	\end{center}
\end{table}

\subsection{Run-time Complexity Analysis}
\label{experiments:complexity}
Moreover, we compare the run-time complexity in the testing stage. Since in general the input LR face has no identity label, GieGAN needs to infer the most possible identity label based on the method described in Sec. \ref{GieGAN} which would consume much computational complexity. For example, for CASIA that contains $10,575$ identities, Table \ref{tab:run-time} shows that GieGAN takes about $61$ s and $227$ s to hallucinate a face from $8\times8$ to $32\times32$ and from $16\times16$ to $64\times64$, respectively. In contrast, both SiGAN and DieGAN are feed-forward networks without the need of estimating the most possible identity label so that they can hallucinate a HR face very quickly. As shown in Table \ref{tab:run-time}, SiGAN, DieGAN, UR-DGN \cite{fhgan02_ultra} and DCGAN \cite{dcgan} takes less than $1$ s to hallucinate a $32\times32$ or $64\times64$ face. In contrast, PRSR \cite{pixelrecurrent}, which is based on a pixel-recurrent structure, needs to predict every pixel during the hallucination, thereby consuming significantly longer time compared to the others. Compared to the other schemes, SiGAN achieves the best visual quality and face recognition/verification rates at a reasonable computational cost, whereas GieGAN achieves comparable visual quality at the cost of high computational complexity due to the need of exhaustive identity label search in the generator. In contrast, the complexity of DieGAN is as low as SiGAN, but it slightly degrades visual quality compared to SiGAN and GieGAN.

\begin{table}
	\caption{\small Run-time complexity comparison in hallucinating one HR faces of SiGAN and the compared methods.}
	\label{tab:run-time}
	\begin{center}
		\begin{tabular}{l|r|r}
			\hline			
			Method & $32\times32$ & $64\times64$  \\
			\hline\hline
			DFCG \cite{song-ijcai17-faceSR}      &   14.24 s  &   21.65 s  \\
			UR-DGN \cite{fhgan02_ultra} &   0.61 s  &   0.89 s  \\
			DCGAN \cite{dcgan}          &   0.55 s  &   0.96 s  \\
			PRSR \cite{pixelrecurrent}  & 227.12 s  & 1091.78 s \\
			GieGAN         	            &  61.12 s  &  227.95 s \\
			DieGAN        	            &   0.57 s  &    0.91 s \\
			SiGAN (proposed)            &   0.71 s  &    0.92 s \\
			\hline
		\end{tabular}
	\end{center}
\end{table}

\subsection{Discussions}
We have presented three identity-embedding GANs for identity-preserving face hallucination: $1)$ SiGAN, $2)$ GieGAN, and $3)$ DieGAN. GieGAN and DieGAN directly embed the identity labels in the training of generator and discriminator, respectively, that requires a fully labeled training set with exact identity labels for all training faces. Since the  generator of GieGAN is directly guided by the identity information during training for identity-preserving reconstruction, it is relatively easy to train, but is computationally very expensive as explained in Sec. \ref{experiments:complexity}. In contrast, the identity information of DieGAN is embedded in  the loss function of the discriminator to authenticate the identities of HR faces hallucinated by the generator, making DieGAN much faster than GieGAN, since it is not required to test all possible identities during hallucination. However, the training of generator in DieGAN is indirectly guided by identity information embedded in the discriminator, making it relatively difficult to train compared to GieGAN, and thereby degrading face hallucination performance. Note, once additional training samples with new identity labels are collected, both GieGAN and DieGAN require a retraining because the number of identities is changed. 

Thanks to its efficient pairwise learning approach, SiGAN can achieve good identity-preserving face hallucination performance at a significantly reduced labeling cost since, without the need of knowing the true identities of faces, it only requires weak pairwise identity labels signifying whether a pair of two faces belong to a same identity. Furthermore, SiGAN can also be easily updated from new training samples, by simply paring the new training samples  with old training faces randomly, and then fine-tuning SiGAN based on the new training pairs, without the need of a retraining. Therefore, SiGAN is a better choice compared to GieGAN and DieGAN.

\section{Conclusion}
\label{sec:conclusion}
We proposed a identity-preserving Siamese face hallucination GAN based on a novel pairwise  learning scheme to capture identity-aware facial representations for reconstructing photo-realistic and identity-preserving HR faces. We have also proposed a new loss function that integrates a reconstruction loss term, a pairwise identity loss term, and a GAN loss term to guide the raining of the proposed GAN to significantly improve the realism of a hallucinated face and its authenticity to the identity. Experimental results demonstrate that our method significantly outperforms state-of-the-art face hallucination networks in terms of objective face recognition/verification rate, while still achieving photo-realistic reconstruction subjectively.



\bibliographystyle{IEEEtran}
\bibliography{face_hallucination}


\end{document}